\documentclass{article}
\usepackage{graphicx} 
\usepackage{geometry}
\usepackage{authblk}
\usepackage{parskip}
\usepackage{booktabs}  
\usepackage{array}     
\usepackage{hyperref}
\usepackage{breakurl}
\usepackage{listings}
\usepackage[table, dvipsnames]{xcolor}
\usepackage{longtable} 
\usepackage{multicol} 
\usepackage{lipsum} 
\usepackage{amsmath} 
\definecolor{mygray}{gray}{0.8}

\usepackage[utf8]{inputenc}
\usepackage[style=apa,sorting=nyt]{biblatex}
\DeclareLanguageMapping{american}{american-apa}
\author[1]{Ekin Ozince}
\author[2]{Yiğit Ihlamur}
\affil[1]{Carnegie Mellon University (Pittsburgh, PA)}
\affil[2]{Vela Partners (San Francisco, CA)}

\title{Automating Venture Capital: Founder assessment using LLM-powered segmentation, feature engineering and automated labeling techniques}
\date{June 2024}

\begin{document}

\maketitle

\begin{abstract}
    This study explores the application of large language models (LLMs) in venture capital (VC) decision-making, focusing on predicting startup success based on founder characteristics. We utilize LLM prompting techniques, like chain-of-thought, to generate features from limited data, then extract insights through statistics and machine learning. Our results reveal potential relationships between certain founder characteristics and success, as well as demonstrate the effectiveness of these characteristics in prediction. This framework for integrating ML techniques and LLMs has vast potential for improving startup success prediction, with important implications for VC firms seeking to optimize their investment strategies.
\end{abstract}

\section{Introduction}
For venture capital (VC) firms, identifying early-stage startups with high potential for success is essential—a difficult task given the countless variables at play. However, as tools like machine learning (ML) evolve, they present key opportunities to supplement VC funds' analyses and improve their investment strategies. Increasingly popular large language models (LLMs) can also support this process by quickly extracting insights from data that is otherwise difficult to parse.

This paper aims to introduce potential pathways for using ML and LLMs to predict the success of new startup founders. We leverage LLMs to propose and apply new features based on limited founder data that reflects the information available when a VC firm considers funding a company. We then use these features to gain insights and make predictions about startup success, demonstrating the vast potential for LLMs in this space.

\section{Methodology}

\subsection{Dataset}
The original dataset comprised two CSV files respectively containing information about 1022 successful and 2694 unsuccessful startup founders. These founders are based in the United States and founded companies in 2010 or later. For the purpose of this paper, "success" means that the founder's company had an initial public offering with a valuation over \$500M, was acquired for over \$500M, or raised over \$500M and is still operating. 

The features of interest in the raw data were \textbf{linkedin\_json} (all information from a founder's LinkedIn profile formatted in JSON) and \textbf{org\_name} (the name of the founder's company). The use of these variables is described further in the following sections. 

To prepare for the prediction of founder success, we combined the two original CSV files into one pandas dataframe and created a boolean flag column called \textbf{success} to distinguish successful and unsuccessful founders. To make the data more manageable for the LLMs, we worked with a random sample containing 150 successful and 150 unsuccessful founders. It should be noted that rows with missing LinkedIn profiles or incorrectly formatted JSON data were deliberately not sampled.

\subsection{Feature engineering}
Engineering features from the raw data involved several steps, including parsing the data through code and using an LLM, specifically OpenAI's GPT-4o, to extract key information.

\subsubsection{Processing education data}
Initially, more traditional techniques were used to extract educational information about each founder from their LinkedIn data. First, we parsed the LinkedIn data to determine each degree that a founder received and used this information to encode a variable, \textbf{highest\_edu}, indicating the highest degree received by the founder. Each education level was mapped to an integer, as shown in Table \ref{tab:degree_mapping}.\footnote{Mapping adapted from \cite{xiong2023}.}

\begin{table}[h!]
\centering
\begin{tabular}{>{\raggedright\arraybackslash}p{8cm} c}  
\toprule
\textbf{Degree} & \textbf{Mapping} \\
\midrule
No degree or unknown & 0 \\
Bachelor's (BA, BS, etc.) & 1 \\
Master's (MA, MS, MBA, etc.) & 2 \\
Doctorate (PhD, MD, JD, etc.) & 3 \\
\bottomrule
\end{tabular}
\caption{Degree mapping}
\label{tab:degree_mapping}
\end{table}

We then parsed the LinkedIn data to determine the fields of study for each founder. Different categories of subjects were also mapped to different integers, as shown in Table \ref{tab:subject_mapping}.\footnote{Mapping adapted from \cite{xiong2023}.} Since one founder could have studied in multiple fields, this information was stored as sets of integer mappings in a variable called \textbf{fields\_of\_study}. Though many founders studied in more than one field, some were assigned empty sets due to limited data about their education. 

\begin{table}[h!]
\centering
\begin{tabular}{>{\raggedright\arraybackslash}p{10cm} c}  
\toprule
\textbf{Fields of study} & \textbf{Mapping} \\
\midrule
Math, quantitative fields, etc. & 0 \\
Biology, chemistry, medicine, psychology, physiology, anatomy, immunology, genetics, etc. & 1 \\
Accounting, banking, actuarial science, finance, economics, etc. & 2 \\
Business, management, entrepreneurship, leadership, etc. & 3 \\
Sales, distribution, marketing, etc. & 4 \\
Computer science/engineering, machine learning, artificial intelligence, HCI, etc. & 5 \\
English, media, film, history, journalism, philosophy, liberal/visual arts, etc. & 6 \\
Political science, sociology, law, consulting, etc. & 7 \\
Architecture, design, urban planning, etc. & 8 \\
Engineering, robotics, physics, etc. & 9 \\
\bottomrule
\end{tabular}
\caption{Fields of study mapping}
\label{tab:subject_mapping}
\end{table}

\subsubsection{Founder summaries}
The rest of the feature engineering process involved significant use of LLMs and prompting techniques, starting with the generation of summaries about each founder. Given a founder's LinkedIn data and latest company name, the LLM was prompted to provide a detailed summary of the founder—including their education, location, skills, and employment history. The prompt emphasized that the model should focus on the time before the founder started their company, as this is the only information that would be available if a venture capitalist was assessing a new founder in a real-life situation.

Here, we used few-shot prompting, wherein the prompt provides examples to guide what the model should output \footnote{\cite{touvron2023}.}. The prompt presented a general structure for the summary, as well as an example summary for a specific founder.\footnote{Prompt adapted from \href{https://github.com/Vela-Engineering/moneyball-founder-segmentation-anushka/blob/main/src/llm_prompting/generate_summary_and_segment.py}{Vela Partners' GitHub.}} We also asked the LLM to search the web for supplementary information, like the statuses of any companies previously started by the founder. See the appendix for the specific prompt used and an instance of the LLM's output.

\subsubsection{Level segmentation}
\textbf{Generating levels} \newline
Previous research\footnote{\cite{wang2024}.} has segmented startup founders into five different categories, from L1 to L5, describing founders' level of experience and achievement in areas related to startup success. To provide more detailed insights, we used an unsupervised LLM-based method to increase the degree of segmentation to ten levels. The LLM was fed fifty founder summaries (generated in the previous section) and asked to propose a list of ten levels building upon the existing list of five. See the appendix for the specific prompt used to propose the ten levels.

We then refined the descriptions of the ten levels to improve specificity and the LLM's ability to produce correct outputs. The levels are as follows:

\begin{itemize}
    \item \textbf{Level 10 (L10)}: Serial entrepreneurs who have successfully founded and exited multiple companies, each with significant funding and valuation, typically above 100M+, and are consistently successful in scaling companies to IPO or large exits.

    \item \textbf{Level 9 (L9)}: Entrepreneurs who have previously led unicorn (valuation $>$ \$1B) startups, before founding their current company.

    \item \textbf{Level 8 (L8)}: Founders known for groundbreaking innovation, who have significantly disrupted or created new markets. Often recognized as thought leaders within the tech industry.

    \item \textbf{Level 7 (L7)}: Founders who have previously scaled startups, securing substantial funding (50M-100M) and paving the way for potential IPO or major exit.

    \item \textbf{Level 6 (L6)}: Founders who have had medium to large-sized exits and may have held key executive roles in notable tech companies. 

    \item \textbf{Level 5 (L5)}: Founders whose previous companies have secured substantial funding (100M+) with pending IPOs.

    \item \textbf{Level 4 (L4)}: Founders with successful small to medium-sized exits or significant roles at notable technology companies.

    \item \textbf{Level 3 (L3)}: Individuals with 10-15 years of technical and management expertise, including significant roles at high-growth tech companies or unicorn startups. May include academia with a PhD in relevant fields.

    \item \textbf{Level 2 (L2)}: Founders with a few years of startup experience prior to founding their company, possibly graduates of well-known accelerator programs.

    \item \textbf{Level 1 (L1)}: Founders who are relatively new to the tech industry, potentially with negligible experience but demonstrating high potential.
\end{itemize}

\textbf{Categorizing founders by level} \newline
After finalizing these levels, we implemented prompt chaining, feeding the previously generated founder summaries into an LLM so that it could categorize each founder as one of these levels. Then, level-based success rates were calculated by determining the number of founders within each level that were deemed to be successful.

This process utilized chain-of-thought (CoT) prompting to build on the few-shot technique.  Few-shot CoT involves providing an example of the input, output, and reasoning that the LLM should follow to produce that output.\footnote{\cite{wei2023}.} See the appendix for the specific prompt used for labelling.\footnote{Prompt adapted from \href{https://github.com/Vela-Engineering/moneyball-founder-segmentation-anushka/blob/main/src/llm_prompting/generate_summary_and_segment.py}{Vela Partners' GitHub.}}

\subsubsection{Persona segmentation}

\textbf{Generating personas} \newline
The process of persona segmentation mimics that of level segmentation.
Prior research\footnote{\cite{suyal2024}.} has worked with ten personas describing key characteristics of founders beyond the information provided by level segmentation. Using a very similar unsupervised method to the previous section, we prompted the LLM to suggest a list of twenty personas based on the existing categories and fifty founder summaries. 

After adjusting the wording for improved results, the final personas are as follows:
\begin{quote}
(A) Successful serial entrepreneur who built a public company.

(B) Successful serial entrepreneur who had a company with exit over \$100M.

(C) Fresh graduate entrepreneur who founded a company immediately or within a year after university.

(D) First-time entrepreneur who founded a company within 3-5 years of finishing university.

(E) Founder holds a PhD degree or is/was a researcher, or is/was a professor at a university.

(F) First-time non-tech entrepreneur with 10+ years of experience in non-tech industries.

(G) Tech executive at tech company/companies for over 10 years.

(H) Engineer/executive at a unicorn company.

(I) Founder who has a background as an engineer or product manager at a tech firm.

(J) Founder with an MBA.

(K) Founder who has lived or worked in multiple countries.

(L) Serial entrepreneur with at least two companies started before.

(M) Founder who came from a corporate background but with less than 10 years of total experience.

(N) Founder with a background in finance, accounting, or economics.

(O) Founder who pivoted career paths from arts/humanities to tech entrepreneurship.

(P) Founder with experience in nonprofit organizations or social entrepreneurship.

(Q) Founder who has received notable awards or recognition in their industry (e.g., Forbes 30 under 30).

(R) Founder with a military service background.

(S) Female founder with notable achievements in male-dominated industries.

(T) Any others that are not in the above categories.
\end{quote}

It is important to acknowledge that the use of unsupervised Large Language Model (LLM) prompting techniques may result in the generation of biased personas, such as (R). For the purposes of this research, we have elected to present the outputs of the unsupervised technique. This transparency aims to provide valuable insights that can inform the development of our future AI modeling techniques with a focus on ethical guardrails in our prompting constructs. 

\textbf{Categorizing founders by persona} \newline
Once again, we implemented prompt chaining and few-shot CoT prompting to generate persona labels for each founder in the dataset.\footnote{Prompt adapted from \href{https://github.com/Vela-Engineering/moneyball-founder-segmentation-anushka/blob/main/src/llm_prompting/generate_summary_and_segment.py}{Vela Partners' GitHub.}} See the appendix for the full prompt. 

In contrast to the level segmentation, founders could be assigned multiple personas. This variable was thus recorded in the data as several boolean columns corresponding to each possible persona. An analysis of success rates by persona followed.

\subsubsection{Additional boolean flags}

\textbf{Generating boolean flags} \newline
To capture further information about startup founders, we added 23 boolean variables to the dataset. These variables were determined through an initial brainstorm and subsequent LLM prompting similar to previous prompting work.

The 23 boolean flags are as follows:
\begin{itemize}
    \item \textbf{is\_firsttime\_founder}: 1 if the founder has created only one company. 0 if the founder has created more than one company.
    \item \textbf{is\_researcher}: 1 if the founder worked as a researcher before starting their company. 0 otherwise.
    \item \textbf{is\_phd}: 1 if the founder received a PhD or similar doctorate degree before starting their company. 0 otherwise.
    \item \textbf{is\_scholarship}: 1 if the founder received a notable scholarship at some point before starting their company (for example, as a Fulbright Scholar). 0 otherwise.
    \item \textbf{is\_young\_grad}: 1 if the founder created their company within a year of finishing university. 0 otherwise.
    \item \textbf{is\_dropout}: 1 if the founder dropped out of university before starting their company. 0 if the founder did not drop out of university.
    \item \textbf{top\_tier\_uni}: 1 if the founder graduated from a top 50 university, according to the QS World University Rankings, before starting their company. 0 otherwise.
    \item \textbf{multiple\_degrees}: 1 if the founder received more than one academic degree before starting their company. 0 otherwise.
    \item \textbf{entrepreneurship\_education}: 1 if the founder studied entrepreneurship or attended an entrepreneurship program (e.g., incubator, accelerator) before starting their company. 0 otherwise.
    \item \textbf{few\_years\_experience}: 1 if the founder had 1-5 years of work experience before starting their company. 0 otherwise.
    \item \textbf{decade\_experience}: 1 if the founder had 10 or more years of work experience before starting their company. 0 if the founder had less than 10 years of experience before starting their company.
    \item \textbf{big\_tech\_experience}: 1 if the founder worked at a big tech company (NVIDIA, Google, Meta, Apple, Amazon, or Microsoft) before starting their company. 0 otherwise.
    \item \textbf{comm\_experience}: 1 if the founder has experience in marketing, public relations, or a related field before starting their company. 0 otherwise.
    \item \textbf{exec\_experience}: 1 if the founder held a senior executive role (e.g., CEO, COO, CTO, VP, President) before starting their company. 0 otherwise.
    \item \textbf{is\_investor}: 1 if the founder worked as an investor (e.g., Venture Capital, Angel Investor) before starting their company. 0 otherwise.
    \item \textbf{is\_board\_member}: 1 if the founder served on a board of directors before starting their company. 0 otherwise.
    \item \textbf{international\_uni}: 1 if the founder attended a university outside the US before starting their company. 0 otherwise.
    \item \textbf{lived\_multiple\_countries}: 1 if the founder lived in multiple countries before starting their company (e.g., studied at universities in multiple countries or worked in different countries before founding their company). 0 otherwise.
    \item \textbf{has\_job\_hopped}: 1 if the founder has changed employers frequently (every 1-2 years) before starting their company. 0 otherwise.
    \item \textbf{has\_promotions}: 1 if the founder received any promotions within a place of work before starting their company (e.g., employment data shows that their role at a certain employer changed over time). 0 otherwise.
    \item \textbf{is\_top\_tier\_consultant}: 1 if the founder worked at a major management consulting firm (e.g., McKinsey \& Company, BCG) before starting their company. 0 otherwise.
    \item \textbf{is\_top\_tier\_banker}: 1 if the founder worked at top tier bank (e.g., Goldman Sachs) before starting their company. 0 otherwise.
    \item \textbf{founded\_under\_30}: 1 if the founder started their company before the age of 30 (use available information like graduation dates to estimate their age when founding). 0 otherwise.
\end{itemize}

\textbf{Applying boolean flags} \newline
The boolean flag values for each founder were generated through few-shot prompting and feeding fifty founder summaries to the LLM. Then, success rates were calculated for each possible value (True or False) of the boolean flags. A shortened version of the prompt used for labelling is in the appendix below.

\subsection{Predicting success through machine learning}
After engineering numeric variables about founders' education, entrepreneurial levels, personas, and more, we ran three machine learning models to classify founder success based on these variables: linear regression, random forest, and XGBoost. The models were trained on a set of 240 instances and tested on two separate test sets to reflect different scenarios that a venture capitalist fund might face. The first test set contained 28 unsuccessful and two successful founders, and the second contained 15 of each group. Performance metrics, such as accuracy and precision, were then used to evaluate and compare the effectiveness of the models.

\section{Results and evaluation}

\subsection{Level segmentation}
The total counts and rounded success rates by level are shown in Table \ref{tab:level_success_rates}. The success rates are below 50 percent for Levels 1 to 4 and above 50 percent for the others. This finding makes sense given that founders assigned higher levels have more experience with entrepreneurship. Levels 5, 9, and 10 have 100-percent success rates, though the data on L5 and L10 is extremely limited. The perfect success rate of L9 (founders with previous unicorn startups), however, is still noteworthy.

\begin{table}[ht]
    \centering
    \begin{tabular}{lcc}
        \toprule
        \textbf{Level} & \textbf{Total count} & \textbf{Success rate (\%)} \\
        \midrule
        L1   & 16  & 37.5 \\
        L2   & 57  & 35.1 \\
        L3   & 58  & 29.3 \\
        L4   & 79  & 40.5 \\
        L5   & 1   & 100 \\
        L6   & 37  & 64.9 \\
        L7   & 14  & 92.9 \\
        L8   & 12  & 91.7 \\
        L9   & 25  & 100 \\
        L10  & 1   & 100 \\
        \bottomrule
    \end{tabular}
    \caption{Success rates by level}
    \label{tab:level_success_rates}
\end{table}

\subsection{Persona segmentation}
Table \ref{tab:persona_success_rates} summarizes the total counts and success rates by persona. Since each founder can be assigned more than one persona, the total counts sum to a number greater than the sample size. Notably, Persona P (founders with experience in nonprofit organizations) has a zero-percent success rate. Meanwhile, Personas Q (founders who have received notable awards) and R (founders with military service background) have success rates of at least 90 percent, indicating potentially desirable traits in startup founders.

\begin{table}[ht]
\centering
\begin{tabular}{lcc}
\toprule
\textbf{Persona} & \textbf{Total count} & \textbf{Success rate (\%)} \\
\midrule
A & 41 & 56.1 \\
B & 21 & 71.4 \\
C & 6 & 50.0 \\
D & 38 & 44.7 \\
E & 56 & 60.7 \\
F & 96 & 43.8 \\
G & 28 & 50.0 \\
H & 19 & 57.9 \\
I & 73 & 63.0 \\
J & 81 & 42.0 \\
K & 31 & 54.8 \\
L & 87 & 51.7 \\
M & 11 & 27.3 \\
N & 52 & 48.1 \\
O & 9 & 22.2 \\
P & 9 & 0.0 \\
Q & 5 & 100.0 \\
R & 10 & 90.0 \\
S & 6 & 16.7 \\
T & 11 & 45.5 \\
\bottomrule
\end{tabular}
\caption{Success rates by persona}
\label{tab:persona_success_rates}
\end{table}

\subsection{Additional boolean flags}
Table \ref{tab:bool_flag_success_rates} shows the total counts and rounded success rates for a subset of the boolean flags. For the sake of brevity, the variables are only displayed if the difference in success rate between True and False values is greater than five percentage points. The greatest differences are seen within \textbf{is\_dropout}, \textbf{is\_top\_tier\_consultant}, and \textbf{is\_top\_tier\_banker}, though there is very little data for True values in each of those variables.

\begin{table}[ht]
\centering
\begin{tabular}{lcc}
\toprule
\textbf{Persona} & \textbf{Total count} & \textbf{Success rate (\%)} \\
\midrule
\rowcolor{mygray}
is\_researcher (No) & 232 & 46.98 \\
\rowcolor{mygray}
is\_researcher (Yes) & 68 & 60.29 \\
is\_phd (No) & 241 & 47.72 \\
is\_phd (Yes) & 59 & 59.32 \\
\rowcolor{mygray}
is\_scholarship (No) & 297 & 49.83 \\
\rowcolor{mygray}
is\_scholarship (Yes) & 3 & 66.67 \\
is\_young\_grad (No) & 295 & 50.17 \\
is\_young\_grad (Yes) & 5 & 40.00 \\
\rowcolor{mygray}
is\_dropout (No) & 298 & 49.66 \\
\rowcolor{mygray}
is\_dropout (Yes) & 2 & 100.00 \\
top\_tier\_uni (No) & 178 & 41.57 \\
top\_tier\_uni (Yes) & 122 & 62.30 \\
\rowcolor{mygray}
entrepreneurship\_education (No) & 280 & 51.07 \\
\rowcolor{mygray}
entrepreneurship\_education (Yes) & 20 & 35.00 \\
few\_years\_experience (No) & 154 & 45.45 \\
few\_years\_experience (Yes) & 146 & 54.79 \\
\rowcolor{mygray}
decade\_experience (No) & 174 & 54.02 \\
\rowcolor{mygray}
decade\_experience (Yes) & 126 & 44.44 \\
big\_tech\_experience (No) & 270 & 47.04 \\
big\_tech\_experience (Yes) & 30 & 76.67 \\
\rowcolor{mygray}
comm\_experience (No) & 246 & 54.88 \\
\rowcolor{mygray}
comm\_experience (Yes) & 54 & 27.78 \\
exec\_experience (No) & 105 & 45.71 \\
exec\_experience (Yes) & 195 & 52.31 \\
\rowcolor{mygray}
is\_investor (No) & 259 & 48.26 \\
\rowcolor{mygray}
is\_investor (Yes) & 41 & 60.98 \\
is\_top\_tier\_consultant (No) & 292 & 49.32 \\
is\_top\_tier\_consultant (Yes) & 8 & 75.00 \\
\rowcolor{mygray}
is\_top\_tier\_banker (No) & 293 & 49.49 \\
\rowcolor{mygray}
is\_top\_tier\_banker (Yes) & 7 & 71.43 \\
founded\_under\_30 (No) & 207 & 47.83 \\
founded\_under\_30 (Yes) & 93 & 54.84 \\
\bottomrule
\end{tabular}
\caption{Success rates by persona}
\label{tab:bool_flag_success_rates}
\end{table}

\subsection{Machine learning models}
Table \ref{tab:model_performance} compares performance metrics for the three models tested. As stated above, test set 1 contains 28 unsuccessful and two successful founders, and test set 2 contains 15 of each group.

Accuracy is simply the proportion of correct predictions among an entire test dataset. Precision measures the proportion of true positives among all positive predictions—that is, the proportion of startups that a VC supports that become successful. The F1 score aims to assess predictive performance by taking the harmonic mean of accuracy and precision. Finally, TPR indicates the True Positive Rate, equal to $\frac{\text{true positives}}{\text{true positives + false negatives}}$. In context, this metric is the proportion of founders that a VC fund chooses out of the total pool of successful founders.  

In general, based on these metrics, linear regression performs better on the dataset with 28 unsuccessful and two successful founders whereas the other models perform better on the evenly-split dataset. The random forest model also performs better than XGBoost overall. 

Among these metrics, precision is likely the most important in the context of venture capital, as one of the main goals of VC funds is to maximize the proportion of successful startups in their portfolio. By this measure, either linear regression or random forest seems to be the best model depending on the makeup of the dataset being used for prediction. When there are few successful founders in the data, linear regression performs best, with a precision of 18.2\%. However, given a more evenly split dataset, random forest is very effective, with a precision of 70.6\%.

\begin{table}[ht]
    \centering
    \begin{tabular}{ccccc}
        \toprule
        \textbf{Model} & \textbf{Accuracy} & \textbf{Precision} & \textbf{F1 score} & \textbf{TPR} \\
        \midrule
        Linear regression (test set 1) & 0.700 & 0.182 & 0.308 & 1.000 \\
        Linear regression (test set 2) & 0.567 & 0.571 & 0.552 & 0.533 \\
        \hline
        Random forest (test set 1) & 0.667 & 0.1 & 0.167 & 0.500 \\
        Random forest (test set 2) & 0.733 & 0.706 & 0.750 & 0.800 \\
        \hline
        XGBoost (test set 1) & 0.667 & 0.1 & 0.167 & 0.500 \\
        XGBoost (test set 2) & 0.667 & 0.647 & 0.688 & 0.733 \\
    \bottomrule
    \end{tabular}
    \caption{Comparing model performance metrics}
    \label{tab:model_performance}
\end{table}

\section{Discussion}
\subsection{Conclusions}
The features engineered for the purpose of predicting startup success—meaning being valued at \$500M in some way—appear very promising. When receiving these features as inputs, all machine learning models explored in this paper perform relatively well, particularly in the context of venture capital and investment. Additionally, the success rates of different founder classifications—such as the high success rate of founders with military backgrounds—suggest notable relationships between certain founder characteristics and their ability to scale a startup.

\subsection{Limitations and future directions}
One limitation in the sampling and prediction process is that founders were considered individually, regardless of whether started their company with a co-founder. Predictions about startup success were made assuming that the given individual was the only founder of the company. However, the characteristics of the entire founding team likely contribute to a startup's success, so our models' predictive ability might be more limited for multi-founder companies.

Another limitation is the reliability of large language models. Although some initial LLM outputs were reviewed manually to ensure correctness and adjust the prompts as needed, not all 300 instances in the sample were inspected in this manner. This somewhat limits the confidence in the results. For instance, there may be inaccuracies in the founders' total years of work experience as outputted by the LLM. The model calculated these figures by summing the durations of a founder's previous jobs, but some jobs may overlap, leading to an overestimation of the number of years in the professional world.

An additional finding about this research is that one of the persona categories, (S), pertains to whether a founder is female. This persona was generated in an unsupervised manner without human intervention. We kept this persona in this research to flag potential issues with LLMs while making decisions in an autonomous way. This is especially important as many firms are introducing autonomous AI agents to augment decision making in their investment decision processes. 

Future research should address these limitations by considering the interactions between the characteristics of co-founders and further refining the prompting strategies. Future research can also examine the importance of certain features in determining startup success. The success rate analysis in this study and prior studies continue to suggest that there are many hidden features that are related to success than others. Therefore, this research topic should be further pursued. 

\newpage
\printbibliography[title={References}]

\newpage
\appendix
\section{Appendix}
\subsection{Founder summaries}
\subsubsection{Prompt}
\begin{lstlisting}
'''
Look at the data delimited in ### below. This data describes a startup founder.

###
{linkedin_json}
###   

The latest founded company of this individual is {org_name}.


Your task is to generate a detailed and accurate summary from it. The summary should be TO THE POINT. The summary should have the following properties -
State the individual's name and current location in the first sentence. Then state the individuals latest founded company, {org_name}, and what year they started this company. Then give a summary of the individual's education. Mention in points - each institution, the degree type, the major, the period of study, and the location of the institution. 
Next, move on to the employment. Here, include a concise description of each job that the individual held BEFORE they founded {org_name}. This description should include the name of the company where they worked, the name of the employer, the job title, the time period (from DD-MM-YYYY to DD-MM-YYYY), the employment duration (for each employment, in years), and the location if available. This description should exclude any jobs that the individual began after they founded {org_name}. Next, highlight the skills of the individual. If this information is missing, use the descriptions field to extract relevant skills. ALL the details mentioned above MUST be included. Give the above output in a paragraph.


In the end, add the three following questions:

1) The universities where the founder studied, along with their global rankings according to the latest QS World University Rankings (university, #ranking):
$Answer1$
2) The total time period of work experience (in years) BEFORE the individual founded {org_name}. For this part, add together the employment durations for the jobs stated earlier.  
$Answer2$
3) The companies founded/co-founded by them BEFORE founding {org_name} in the format - company, status, net worth in USD. (search on the web to find the exact figures of the company's net worth.):
$Answer3$


Answer the questions in the format according to provided information. In question 3, exclude {org_name}, as this is the latest founded company. Also in question 4, fields including status and net worth in USD should be found in the provided employers' information.
   Represent thousands as "K" and millions as "M" for financial figures, keeping only one digit after the decimal point. For example, use "20.5K" for 20,500 USD and "50.7M" for 50,700,000 USD.
  
Here is an example:
   XXXXX, a male currently residing in San Diego, California, USA. His latest founded company is 10x Genomics, which he co-founded in 2012. He has an extensive educational background, including a PhD in Bioengineering and Biomedical Engineering from the University of California San Diego, where he also completed a Master's degree in Bioengineering and a Bachelor's degree in Bioengineering. In total, he has had 3 educational experiences with a cumulative duration of 14 years. In his professional career, before founding 10x Genomics, XXXXX served as a Research Scientist at Lawrence Livermore National Lab for 3.92 years (2005-2008), as a Biophysicist at Lawrence Livermore National Laboratory for 6 years (2002-2008), and as a Sr Principal Scientist at Bio-Rad Laboratories, Inc. for 0.5 years (2011-2012). He co-founded Quantalife and served as Chief Scientific Officer for 3.16 years (2008-2011). His skills range from muscle mechanics to startup management and include expertise in biomedical engineering, molecular biology, and leadership. Tangney has also authored numerous articles related to cardiac mechanics and myocardial function.

    1) The universities studied, along with their global rankings according to the latest QS World University Rankings (university, #ranking):
    University of California San Diego, #53.

    2) The total time period of work experience (in years) BEFORE the individual founded {org_name}. For this part, add together the employment durations for the jobs stated earlier.
    13.58 years.

    3) The companies founded/co-founded by them BEFORE founding 10x Genomics in the format - company, status, net worth in USD. (search on the web to find the exact figures of the company's net worth.):
    Quantalife, closed, 30.1M.
      '''
\end{lstlisting}

\subsubsection{Example output}
\begin{lstlisting}
'''
XXXXX currently resides in San Francisco, California, USA. He co-founded Airtable in June 2012. He has an extensive educational background, having graduated from Duke University with a BSE in Electrical and Computer Engineering, Economics, studying there from 2004 to 2008.

Before founding Airtable, XXXXX held several notable positions:
1. **Google**: Product Manager from 2009 to 2012. He led a team for the new Google Maps interface and handled imagery and 3D interaction, as well as managing social applications on Android. Duration: 3 years.
2. **Accenture**: Technology Labs Development Group member from August 2008 to August 2009. He designed and built prototypes, including an enterprise iPhone app and a web optimization/analytics platform. Duration: 1 year.
3. **Scripps Institution of Oceanography**: Engineering Research Intern from May 2006 to August 2006. He worked on various engineering research projects. Duration: 0.33 years (approximately 4 months).

Based on the job descriptions, it can be inferred that XXXXX has skills in product management, software development, social applications (i.e., Facebook and Twitter integration), enterprise application design, web optimization, and 3D interaction within software applications.

1) The universities where the founder studied, along with their global rankings according to the latest QS World University Rankings (university, #ranking):
   Duke University, #50.

2) The total time period of work experience (in years) BEFORE the individual founded Airtable. For this part, add together the employment durations for the jobs stated earlier:
   4.33 years.

3) The companies founded/co-founded by them BEFORE founding Airtable in the format - company, status, net worth in USD. (search on the web to find the exact figures of the company's net worth.):
   Not available.
'''
\end{lstlisting}

\subsection{Level segmentation prompts}
\subsubsection{Proposing new levels}
\begin{lstlisting}
'''
  Look at the list delimited in ### below.

  ###
  {summaries}
  ###

  This is a list of strings, separated by commas, where each string is a description of a startup founder. We want to categorize each founder into one of ten different categories. Currently, we only have five categories, delimited in $$$ below. We will build upon these throughout the task.

  $$$
              Level 5 (L5): Entrepreneur whose company has 100M+ funding is going through IPO.
              Level 4 (L4): Entrepreneur who has had a small to medium-size exit or has worked as an executive at a notable technology company
              Level 3 (L3): 10-15 years of technical and management experience (e.g. working at big tech and unicorn startups or having a PhD)
              Level 2 (L2): Entrepreneurs with a few years of experience or accelerator graduates
              Level 1 (L1): Entrepreneurs that are disconnected from tech circles or that have negligible experience but with large potential
  $$$

  Your task is to analyze the list of descriptions delimited in ### to propose ten categories that can be used to classify the founders based on their work and entrepreneurial experience. The ten categories should be similar to, but expand upon, the existing categories given in $$$. 
  '''
\end{lstlisting}

\subsubsection{Labelling founders by level}
\begin{lstlisting}
'''
       Properly understand and analyze the classification of levels given in ###. This will be used in the segmentation of the founder throughout the task.
        ###
Level 10 (L10)
- **Serial Entrepreneur with Multiple 100M+ Exits:** Entrepreneurs who have successfully founded and exited multiple companies, each with significant funding and valuation, typically above 100M+, and are consistently successful in scaling companies to IPO or large exits.
- **Example:** Founder who has taken multiple companies public or sold them for hundreds of millions.

Level 9 (L9)
- **Founder with Previous Unicorn Startups:** Entrepreneurs who have previously led unicorn (valuation > $1B) startups, before founding their current company.
- **Example:** Former CEO of a high-profile startup valued over a billion dollars.

Level 8 (L8)
- **Tech Visionaries and Industry Innovators:** Founders known for groundbreaking innovation, who have significantly disrupted or created new markets. Often recognized as thought leaders within the tech industry.
- **Example:** Founders with revolutionary products that have reshaped industries (e.g., Elon Musk-level innovation).

Level 7 (L7)
- **High-Growth Scale-Up Entrepreneurs:** Founders who have previously scaled startups from seed stages to mid/late stages, securing substantial funding (50M-100M) and paving the way for potential IPO or major exit for companies prior to their current one.
- **Example:** Entrepreneurs who previously ran rapidly scaling mid-sized startups.

Level 6 (L6)
- **Prominent Executives with Solid Exit Histories:** Entrepreneurs who have had medium to large-sized exits and may have held key executive roles in notable tech companies. Typically, these founders have substantial industry experience.
- **Example:** Former executives at notable tech companies who have sold their own startups.

Level 5 (L5)
- **Growth-Stage Entrepreneurs with Pending IPOs:** Entrepreneurs whose previous companies have secured substantial funding (100M+) and are in the process of going through an IPO.
- **Example:** Former CEO of a company on the brink of public offering.

Level 4 (L4)
- **Experienced Founders in Mid-Sized Exit and Notable Roles:** Entrepreneurs with successful small to medium-sized exits or significant roles at notable technology companies, often recognized for solid contributions to previous ventures.
- **Example:** Entrepreneur with a $50M+ exit or former senior executive at a notable tech firm.

Level 3 (L3)
- **Technical Experts with Deep Industry Experience:** Individuals with 10-15 years of technical and management expertise, including significant roles at high-growth tech companies or unicorn startups. Also may include academia with a PhD in relevant fields.
- **Example:** Senior engineer or technical lead at a unicorn startup, or PhD with relevant industry patents.

Level 2 (L2)
- **Early-Stage Entrepreneurs or Accelerator Alums:** Entrepreneurs with a few years of startup experience prior to founding their company, possibly graduates of well-known accelerator programs, and showing early promise in their ventures.
- **Example:** Founder who has gone through Y Combinator or another renowned accelerator.

Level 1 (L1)
- **Nascent Founders with High Potential:** Entrepreneurs who are relatively new to the tech industry, potentially with negligible experience but demonstrating high potential and ambition.
- **Example:** Recent graduates or budding entrepreneurs with innovative ideas but limited track record.
###

Imagine a committee of 3 experienced Venture Capital (VC) analysts, who will review a detailed summary of the founder's background and accomplishments. The steps are as follows:
The analysts will examine the data and independently note down any details that might play a crucial role in the final evaluation and level allotment.
Each analyst, using the collected data and their specific expertise, will independently assign a founder level that they believe best fits the founder's qualifications, giving clear justifications for their decisions. These justifications MUST be supported by facts and figures to ensure they are quantitative and precise.
The analysts will then share their level assignment with each other. If all 3 analysts have the same assignment, this is the final one. If the assignments differ, each analyst will take turns to explain their reasoning to the others. This will continue till everyone agrees on one level.

To avoid hallucination, don't be influenced by previous queries.

An overall report will be generated containing the analysis from each analyst, the assigned level, and the justifications for this assignment.
Example -
Using the provided founder summary as an example, here are potential 'Analyst Observations and Level Assignments':

Founder Summary:
```
XXXXX, a 40-year-old male, currently resides in El Segundo, California, USA. He had his education at UCLA Anderson School of Management, where he obtained a Master's degree for a 6-year program between January 2007 and January 2009. His Bachelor's degree, a 4-year program, was awarded by Princeton University, studied between January 2000 and June 2004. His latest founded company is Vulcury, which he founded in 2022. His professional career spans across multiple roles and industries. Before founding Vulcury, he began as a Venture Advisor at Soundproof Ventures from May 2021 to Present. He was CEO at 3Diligent from July 2014 until August 2022 (8 years). Before founding 3Diligent, he worked as Manager - Strategy at Deloitte Consulting from September 2012 to July 2014 (2 years), Senior Consultant at Deloitte from June 2009 to September 2012 (3 years), Founder at Citi Habitats Graduate Relocation Program from April 2005 to April 2010 (5 years), and Director of Marketing and Operations at Winston Churchill High School from January 2007 to January 2010 (3 years). XXXXX is equipped with a multitude of skills, which include but not are limited to supply chain management, internet marketing, advertising, management, economics, marketing, teaching and sales.

1) The universities studied, along with their global rankings according to the latest QS World University Rankings (university, #ranking):
$UCLA Anderson School of Management, #16$
$Princeton University, #13$
2) The total time period of work experience (in years) BEFORE the individual founded 10x Genomics.
$18 years.$
3) The companies founded before the latest founded company (company, status, total funding in USD):
$3Diligent, Active, N/A$
```

Analyst Observations and Level Assignments:
Analyst 1:
Observation: XXXXX has 20 years of prior work experience, has held leadership positions at multiple companies, and has a Master's degree from a renowned university. However, there's no evidence of founding a business with funding over $10 million.
Level Assignment: L3 - Justification: The sum of his experience and education align him well with L3, which requires 10-15 years of technical and management experience.

Analyst 2:
Observation: XXXXX has formed a company (3Diligent), held several high-level roles, and his duration of work experience exceeds the 10-15 years of experience requirement stated in L3.
Level Assignment: L3 - Justification: While he has significant experience in several high-profile roles, there's no evidence of his companies achieving extremely high funding or going through an IPO, excluding him from L4 or above.

Analyst 3:
Observation: XXXXX's education background includes a Master's degree from the #16 ranked UCLA Anderson School of Management and a Bachelor's from the #13 ranked Princeton University.
Level Assignment: L3 - Justification: His degrees from top rankings institutes and his varied work experience qualify him for L3. However, the lack of high funding or IPO prevents assignment to L4 or above.

In this example, all 3 analysts independently agreed on assigning Level 3 to the founder XXXXX. The key justification centers around his vast experience in different roles and businesses he has founded, but the absence of evidence about his companies securing high levels of funding or going through an IPO, parameters required for Levels 4 and above. Therefore, the final decision, supported by factual and quantitative justifications, designates XXXXX as Level 3.

Conduct a similar analysis for the founder summary provided below. The output should ONLY be the level assignment, example "L10".
```
{summary}
```
'''
\end{lstlisting}

\subsection{Persona segmentation prompt}
    

\subsubsection{Labelling founders by persona}
\begin{lstlisting}
'''
       The prompt for using founder summary to classify the founder into one of the 20 personas (A to T) is as follows:

Properly understand and analyze the classification of personas given in ###. This will be used in the segmentation of the founder throughout the task.

###
(A) Successful Serial Entrepreneur who built a public company.
(B) Successful Serial Entrepreneur who had a company with exit over $100M (search up the funding of the companies founded by them on google to get exact figures).
(C) Fresh Graduate Entrepreneur who founded a company immediately or within a year after university. (Lookup the summary provided and find the difference between the year of finishing studies and the year of founding the company. Do not select this persona if there is no information about university graduation date.).
(D) First time entrepreneur who founded a company within 3-5 years of finishing university (Look up the summary provided and find the difference between the year of finishing studies and the year of founding the company. Do not select this persona if there is no information about university graduation date.).
(E) Founder holds a PhD degree or is/was a researcher, or is/was a professor at a University.
(F) First Time Non-tech Entrepreneur who has 10+ years of experience in non-tech industries.
(G) Tech Executive at tech company/companies for over 10 years (use employment durations to sum the total work experience in tech firms).
(H) Engineer/executive at a unicorn company (use top 1000 Unicorn companies data available on the internet).
(I) Founder who has a background as an engineer or product manager at a tech firm.
(J) Founder with an MBA.
(K) Entrepreneur who has lived or worked in multiple countries.
(L) Serial Entrepreneur with at least two companies started before.
(M) Founder who came from a corporate background but with less than 10 years of total experience.
(N) Founder with a background in finance, accounting, or economics.
(O) Founder who pivoted career paths from arts/humanities to tech entrepreneurship.
(P) Founder with experience in nonprofit organizations or social entrepreneurship.
(Q) Founder who has received notable awards or recognition in their industry (e.g., Forbes 30 under 30).
(R) Founder with military service background.
(S) Female founder with notable achievements in male-dominated industries (assume a founder is female if the summary includes she/her pronouns and assume a founder is male if the summary includes he/him pronouns).
(T) Any others that are not in the above categories.
###

Try to avoid classifying as (T) unless the founder fulfils none of the other criteria.
At the same time, don't be overly optimistic while classifying. Your classification must purely be based on facts. Each founder must be assigned to at least one class. They can be assigned to more than one class if multiple are fitting. 

To avoid hallucination, don't be influenced by previous queries.

Delimited below in $$$ is an example of a founder summary and its analysis.
$$$
Here is the summary:    

XXXXX, a male currently residing in Beijing, China. He holds a Bachelor's degree in Information Engineering from Cambridge University (2003-2007), a Master's degree in Study Information Engineering from Cambridge University (2003-2007), a Bachelor's degree in Electrical and Information Engineering from MIT (2005-2006), and completed High School at Millfield (2001-2003). XXXXX most recently founded 17zuoye (2011-present) in Beijing, China in 2011. Before founding 17zuoye, he was a Co-Founder at D&H Ltd. (2009-2011) in London, United Kingdom and Beijing, China; Associate Analyst at UBS (2007-2009) in Zurich, Switzerland; and Expert Assistant at Beijing Organising Committee for Olympic Games (2007-2008). His total work experience before founding 17zuoye was 4 years. XXXXX's skills span across financial services, stock trading, presentation, teaching, marketing, and more.

1) The universities studied, along with their global rankings according to the latest QS World University Rankings (university, #ranking):
    Cambridge, #7;
    MIT, #1.

2) The total time period of work experience (in years) BEFORE the individual founded 10x Genomics.
    4 years.

3) The companies founded before the latest founded company (company, status, total funding in USD):
    D&H Ltd., closed, 20.5M.


Here is the analysis that helps arrive to a final labelling for the founder:

Founded D&H Ltd. in 2009 (before founding 17zuoye): He is not a first-time entrepreneur, so he does not meet criterion (D) or (F). Both his companies were founded more than a year after he graduated university, so he does not meet criterion (C). Since he has founded two companies, he is a serial entrepreneur. However, the exit details or current valuation for the companies aren't available, so we cannot categorize him under (A) or (B). Also, he started only one company before founding his current company, 17zuoye, so he does not fit persona (L).  He does not hold a PhD degree nor was a researcher or professor: does not meet criterion (E). He has not been a tech executive for over 10 years: does not match persona (G). The given information does not indicate that he has worked at a unicorn company, disqualifying him for persona (H). He has not worked as an engineer or product manager at a tech firm, so he does not fit persona (I). He does not have an MBA, so he does not meet criterion (J). He has lived in multiple countries throughout his career, including China and the United Kingdom, fitting the immigrant persona (K). Based on his employment history, he had less than 10 years of experience in a corporate background, as well as experience in finance (UBS is a bank) before founding 17zuoye; therefore, he fits personas (M) and (N). He does not have experience in the arts/humanities, nonprofits, or military service, disqualifying him from personas (O), (P), (R). He has not received notable industry recognition, so he does not fit persona (Q). Finally, he is not female, so he does not fit persona (S). 

Based on this analysis, we can categorize XXXXX as personas (K), (M), (N).

Final assignment: [K, M, N]
$$$           

Conduct such an analysis for the founder summary below. The output should ONLY be the level assignment(s) in a list format, example "[D, J]". There should be no other output apart from this list.

Summary to analyze:    
```
{summary}
```
'''
\end{lstlisting}

\subsection{Boolean flag prompts}

\subsubsection{Labelling boolean flags}
\begin{lstlisting}
'''
       Properly understand and analyze the boolean variables given in ### that each describe characteristics of startup founders. This will be used in the classification of the founder throughout the task.

###
is_firsttime_founder: 1 if the founder has created only one company. 0 if the founder has created more than one company.

is_researcher: 1 if the founder worked as a researcher before starting their company. 0 otherwise.

is_phd: 1 if the founder received a PhD or similar doctorate degree before starting their company. 0 otherwise.
###

Your task will be to assign a value to each of these variables based on a written summary about a founder. Each variable can only have one value, either 0 or 1. To avoid hallucination, don't be influenced by previous queries.

Here is an example:

***Founder summary:***
```
XXXXX, currently residing in the Los Angeles Metropolitan Area, California, USA. His latest founded company is Applariat, which he started in March 2016. He attended California State Polytechnic University-Pomona, where he earned a Bachelor of Science (BS) degree and a Bachelor of Engineering (BE) degree in Mechanical Engineering, both spanning from 1987 to 1993.

Before he founded Applariat, XXXXX held several roles:
- **Michelle's Place (Board of Directors)**: Nov 2001 - Present, 21.92 years.
- **VMWare (Consulting Architect and Practice Manager, Customer Oriented R&D)**: Sep 2010 - Dec 2015, 5.25 years.
- **Integrien Corporation (Senior Director of Professional Services)**: Sep 2007 - Aug 2010, 2.92 years.
- **TIBCO Software (Engagement Manager)**: Apr 2004 - Sep 2007, 3.42 years.
- **EDS (Senior Consultant)**: Jul 1994 - Apr 2004, 9.75 years.

XXXXX's skills include Customer Delivery and Program Management, Service Oriented Architecture, Business Process Management, Infrastructure and Application Operations and Management, Machine Learning and Analytics, and Non-Profit Management. 

1) The universities where the founder studied, along with their global rankings according to the latest QS World University Rankings (university, #ranking):
   - California State Polytechnic University-Pomona, #N/A.
   
2) The total time period of work experience (in years) BEFORE the individual founded Applariat. For this part, add together the employment durations for the jobs stated earlier.
   - 21.34 years.

3) The companies founded/co-founded by them BEFORE founding Applariat in the format - company, status, net worth in USD:
   - Michelle's Place, active, unknown. 
```

***Boolean flag variable assignments:***
```
is_firsttime_founder = 0
is_researcher = 0
is_phd = 0
```

***Final output:***
[0,0,0]

Conduct a similar analysis for the founder summary provided below, and output an integer list containing the value for each variable (a list of length 16 containing 0s and 1s). The output should ONLY be the integer list.

***Founder summary:***
```
{summary}
```
      '''
\end{lstlisting}

\end{document}